\documentclass[times,review,10pt]{elsarticle}

\usepackage{amsmath,amssymb,bm}

\usepackage{graphicx}
\usepackage{booktabs,longtable,tabularx,threeparttable,multirow,array,makecell}

\usepackage[linesnumbered,ruled]{algorithm2e}

\usepackage{xcolor}
\usepackage{tcolorbox}
\tcbuselibrary{breakable,skins}
\definecolor{lightgreen}{RGB}{189,231,189}
\definecolor{lightred}{RGB}{255,182,179}

\usepackage{subfigure}
\usepackage{caption}

\usepackage{url,doi}
\usepackage{xurl}
\usepackage{hyperref}
\hypersetup{
  colorlinks=true,
  linkcolor=blue,
  citecolor=blue,
  urlcolor=black
}
\usepackage[capitalise]{cleveref}

\usepackage{setspace}
\usepackage{lineno}
\usepackage{microtype}

\begin{document}

\begin{frontmatter}

\author[inst1]{Danyu Li}
\ead{3230002471@student.must.edu.mo}

\author[inst1]{Ling Zhou}
\ead{lzhou@must.edu.mo}

\author[inst1,inst2]{Rubing Huang\corref{cor1}}
\ead{rbhuang@must.edu.mo}
\cortext[cor1]{Corresponding author.}

\author[inst5]{Xian Zhong}
\ead{zhongx@whut.edu.cn}

\author[inst3]{Bin Zou}
\ead{binzou2009@ujs.edu.cn}

\author[inst6]{Kui Jiang}
\ead{jiangkui@hit.edu.cn}

\affiliation[inst1]{organization={School of Computer Science and Engineering, Macau University of Science and Technology},
            city={Macao SAR},
            postcode={999078},
            country={China}}

\affiliation[inst2]{organization={Macau University of Science and Technology Zhuhai Research Institute},
            city={Zhuhai},
            state={Guangdong},
            postcode={519099},
            country={China}}

\affiliation[inst5]{organization={Hubei Key Laboratory of Transportation Internet of Things, School of Computer Science and Artificial Intelligence, Wuhan University of Technology},
            city={Wuhan},
            state={Hubei},
            postcode={430070},
            country={China}}

\affiliation[inst3]{organization={School of Food and Biological Engineering, Jiangsu University},
            city={Zhenjiang},
            state={Jiangsu},
            postcode={212013},
            country={China}}

\affiliation[inst6]{organization={School of Computer Science and Technology, Harbin Institute of Technology},
            city={Harbin},
            state={Heilongjiang},
            postcode={150001},
            country={China}}           
            

\title{EGRL: Edge generation-guided relation-aware learning for RNA-protein interaction prediction}

\makeatother

\begin{abstract}
\textit{RNA-Protein Interactions} (RPIs) are critical for regulating cellular functions. While traditional wet-lab experiments for RPI detection are costly and time-consuming, \textit{Deep Learning} (DL) methods provide an efficient computational alternative for \textit{RPI Prediction} (RPIP). In particular, \textit{Graph Neural Networks} (GNNs) are promising, as they naturally model RPI networks. However, existing GNN-based methods often rely on homogeneous graphs or predefined meta-paths, which limit their ability to handle data sparsity and to generalize to cold-start scenarios involving unknown molecules. To address these limitations, we propose \underline{\textbf{E}}dge \underline{\textbf{G}}eneration-guided \underline{\textbf{R}}elation-aware \underline{\textbf{L}}earning (EGRL), a novel framework with several key components: implicit meta-path learning to capture relational semantics without handcrafted paths; a multi-relation-aware attention mechanism for adaptive fusion of interaction patterns; a graph generator that predicts potential (``soft'') edges to support cold-start nodes; and a multi-feature fusion predictor for final interaction scoring. EGRL is jointly trained with a primary task loss and an auxiliary generator loss. Comprehensive evaluations on four benchmark datasets demonstrate that EGRL achieves competitive overall performance. More importantly, it exhibits superior generalization in cold-start settings, achieving an \textit{Area Under the Receiver Operating Characteristic curve} (AUROC) of 0.867 and an \textit{Area Under the Precision-Recall curve} (AUPR) of 0.861 on unknown molecules, corresponding to improvements of 8.6\% in AUROC and 5.0\% in AUPR over prior state-of-the-art methods. The code will be released soon.

\end{abstract}


\end{frontmatter}

\section{Introduction}

The \textit{RNA-Protein Interactions} (RPIs) play a vital role in various cellular processes, including gene regulation~\cite{avila2024exploring} and protein synthesis~\cite{wanowska2022role}, with significant implications for disease prediction~\cite{zhang2026improving}, functional annotation of biomolecules~\cite{xu2022building}, and drug design~\cite{bertoldo2023rna,XIE2026113556}. Traditional biological experiments for RPI detection are often expensive and time-consuming~\cite{stenum2023rna}, thereby motivating the development of computational approaches.

Early computational methods for \textit{RPI Prediction} (RPIP) mainly relied on traditional machine learning models, such as support vector machines and random forests, using hand-crafted features extracted from sequence or structure information~\cite{han2004prediction,liu2010prediction}. Although these methods achieved some success, their performance was limited by the quality of manual feature engineering and their inability to capture complex nonlinear interaction patterns. Subsequently, \textit{Deep Learning} (DL) has gained increasing attention in RPIP due to its ability to automatically learn complex features and interaction patterns from biological data~\cite{li2025ribonucleic}. DL techniques, especially \textit{Convolutional Neural Networks} (CNNs) and \textit{Recurrent Neural Networks} (RNNs), have been widely adopted for RPIP. CNNs extract local motif patterns from sequences, while RNNs and their variants model sequential dependencies. However, these methods typically treat each RNA-protein pair independently and ignore the global relational structure among multiple RNAs and proteins~\cite{patiyal2023prediction, pan2018prediction}.

In recent years, \textit{Graph Neural Networks} (GNNs) have emerged as a powerful framework for modeling RPIs. GNNs naturally represent biological entities as nodes and their interactions as edges in a graph, enabling the modeling of relational structure and dependencies within complex biological systems. By recursively aggregating information from neighboring nodes, GNNs learn enriched node representations that encode both intrinsic features and topological context. This capability makes them well-suited for analyzing diverse biological networks, including RNA-protein interaction graphs~\cite{shen2023prediction}, protein-protein interaction graphs~\cite{yang2024end,BI2026112563}, and RNA-RNA interaction graphs~\cite{yin2025autonomously,WEI2026112078}.

However, conventional GNN-based methods still face notable limitations in RPIP. Many approaches rely on homogeneous graph constructions~\cite{zhang2022deeppn,zhuo2022predicting}, which oversimplify the inherent heterogeneity of biological systems, where nodes (\textit{e.g.}, RNAs and proteins) and interactions (\textit{e.g.}, binding and functional association) are of distinct types. Although some studies incorporate predefined meta-paths to model higher-order relations, such designs require substantial domain expertise and may fail to capture all relevant interaction patterns~\cite{liang2023hmcda,liu2024meta}. These limitations restrict model expressiveness, hinder generalization under sparse data, and limit performance in predicting interactions involving uncharacterized molecules (i.e., the cold-start problem).

To address these challenges, we propose \underline{\textbf{E}}dge \underline{\textbf{G}}eneration-guided \underline{\textbf{R}}elation-aware \underline{\textbf{L}}earning (EGRL), a novel heterogeneous GNN framework that integrates implicit meta-path modeling, a graph generator for cold-start nodes, and multi-relational \textit{Graph Attention Network} (GAT) layers with multi-feature fusion. The main contributions are summarized as follows:
\textit{(1) Implicit meta-path learning:} We propose an implicit meta-path learner that automatically captures the importance of different relation types without manual design, dynamically integrating relation-level semantics into node representations.
\textit{(2) Graph generator for cold-start nodes:} We introduce a jointly trained graph generator that predicts soft edges for unseen RNAs/proteins based on their sequence features, enabling effective cold-start interaction prediction.
\textit{(3) Multi-relational GAT with multi-feature fusion:} We design a multi-relational GAT that processes each edge type independently, along with a predictor that combines concatenation, element-wise product, and absolute difference of node embeddings for enhanced interaction modeling.
\textit{(4) Competitive performance and effective cold-start generalization:} Extensive experiments on four benchmark datasets demonstrate that EGRL achieves competitive performance against state-of-the-art methods and exhibits strong generalization under molecule hold-out and sequence-cluster-based cold-start settings.

The remainder of this paper is organized as follows. \cref{SEC:Related Work} reviews related work; \cref{SEC:Motivation} presents the research motivation; \cref{SEC:Methodology} describes the proposed EGRL framework; \cref{SEC:Experiment} details the experimental setup; \cref{SEC:Results} reports and analyzes the results; and \cref{SEC:Conclusion and Future Work} concludes the paper and discusses future directions.

\begin{figure*}[!t]
    \centering
    \includegraphics[width = \textwidth]{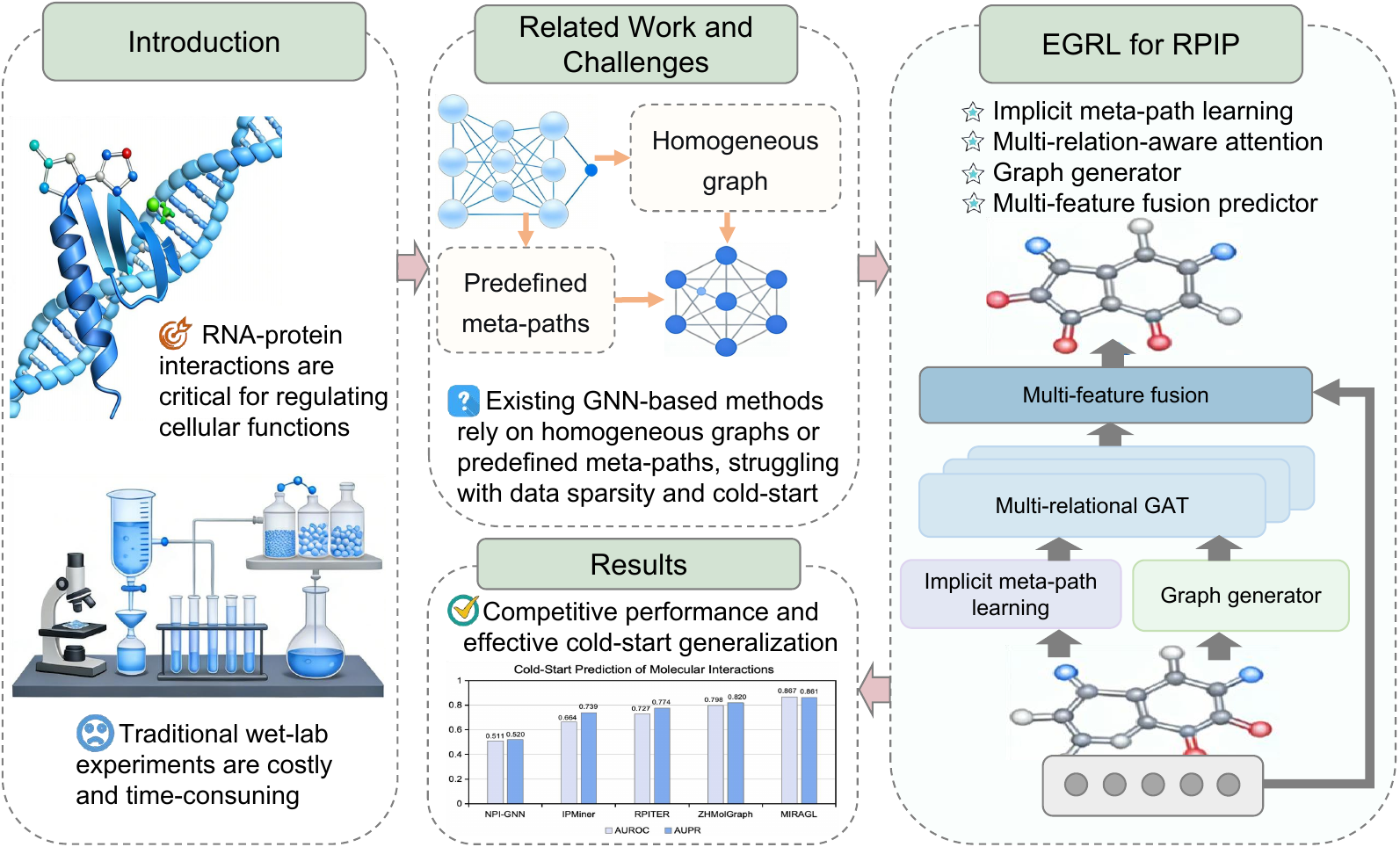}
    \caption{\textbf{Graphical abstract of EGRL.} Overview of the proposed framework and its key components for RNA-protein interaction prediction.}
    \label{FIG:graphical_abstract}
\end{figure*}

Figure~\ref{FIG:graphical_abstract} presents the graphical overview. It highlights the importance and challenges of RPIP, emphasizing the limitations of traditional GNNs based on homogeneous graphs or predefined meta-paths under sparse and cold-start conditions. It further illustrates how EGRL addresses these challenges, achieving competitive overall performance while maintaining strong generalization in cold-start scenarios.

\section{Related work}
\label{SEC:Related Work}

Traditional machine learning methods for RPIP typically rely on manually extracted features. For example, RPI-Pred~\cite{suresh2015rpi} utilizes sequence and structural information with a support vector machine classifier. Despite their interpretability and low computational cost, these methods suffer from limited expressive power, as hand-crafted features may fail to capture complex interaction patterns.

Deep learning approaches have significantly advanced RPIP by automatically learning hierarchical representations. CNNs extract local motif patterns from RNA or protein sequences treated as one-dimensional signals. For instance, MCNN~\cite{pan2022mcnn} predicts RNA-protein binding sites by integrating multiple CNNs, each processing RNA sequence segments of different lengths. RNNs, particularly long short-term memory (LSTM) models, capture long-range dependencies. Some methods further combine CNNs and RNNs to leverage both local and sequential patterns~\cite{zhang2019deepdrbp,pan2018prediction}. In addition, RPI-CapsuleGAN~\cite{WANG2023109626}, a generative adversarial network-based model, employs adversarial training to learn data distributions and enhance feature representations for RPIP.

In recent years, \textit{Graph Neural Networks} (GNNs) have become a mainstream approach for RPIP and have achieved remarkable performance. The GNN paradigm aligns naturally with RNA-protein interaction modeling, where RNAs and proteins are represented as nodes and interactions as edges. Consequently, RPIP can be formulated as a link prediction problem. GNNs learn node representations by aggregating neighborhood information, enabling effective prediction of potential interactions. Several GNN-based methods have been proposed for RPIP. LPICGAE~\cite{zhao2023predicting} incorporates graph autoencoders to infer potential interactions. LncPNet~\cite{zhao2022predicting} designs a network embedding approach for lncRNA-protein interaction prediction. RPI-GGCN~\cite{wang2024rpi} integrates gated graph convolutional networks with a co-regularized variational autoencoder, leveraging Gated Recurrent Units and \textit{Graph Convolutional Networks} (GCNs) to extract topological information. DeepPN~\cite{zhang2022deeppn} employs parallel CNN and GCN layers to capture hidden features of RNA sequences for RPIP. LPI-KCGCN~\cite{shen2023prediction} utilizes a two-layer GCN to learn latent representations of RNAs and proteins and subsequently predicts their interaction scores.

However, many GNN-based methods treat the RPI graph as homogeneous and primarily focus on enhanced node feature learning, which oversimplifies the intrinsic heterogeneity of biological systems. In practice, RPI graphs are heterogeneous, involving multiple types of nodes and interactions. To address this, heterogeneous graph learning methods have been explored~\cite{li2024heterogeneous}. For example, BiHo-GNN~\cite{ma2023predicting} integrates homogeneous and heterogeneous network features through bipartite graph embedding, coupled with a mutual optimization strategy. Meta-paths have also been widely adopted to capture higher-order semantic relations in heterogeneous graphs, such as in RNA-disease association prediction~\cite{deng2020predicting}. An explicit multilevel meta-path aggregation graph embedding model has been proposed for miRNA-disease association prediction~\cite{cao2023metapath}. These studies suggest that meta-path-based feature learning is a viable strategy for modeling heterogeneous graph structures.

\textit{Graph Attention Networks} (GATs) have been further introduced to enhance representation learning in heterogeneous interaction graphs. GATs adaptively aggregate information from neighboring nodes by learning attention coefficients, and have been shown to outperform conventional sequence-based models such as CNNs and bidirectional LSTM networks~\cite{wang2022self}. NABind~\cite{jiang2023structure} integrates sequence and structural descriptors with an attention mechanism to aggregate edge information. RPIembeddor~\cite{matus2024rna} employs a GAT-based framework to incorporate structural and functional features for RPI classification. Unlike self-attention in Transformers, which relies on positional encoding, graph attention computes importance based on node features and graph connectivity, making it more suitable for graph-structured data. In RPIP, modeling the importance of interaction patterns between nodes is more meaningful than relying on absolute positional information. Therefore, GAT provides a natural and effective framework for RPI modeling.

Based on the above observations, we propose EGRL. To the best of our knowledge, this is the first study that integrates implicit meta-path learning, a graph generator for cold-start nodes, and a multi-relational GAT with multi-feature fusion for RPIP.

\section{Motivation}
\label{SEC:Motivation}

RPIs constitute core regulatory mechanisms of cellular activities and are closely associated with the occurrence and progression of various diseases. Accurate RPI prediction is therefore of great significance for drug target discovery, disease diagnosis, and therapeutic development. In recent years, deep learning models have substantially improved the efficiency of RPIP; however, several key challenges remain:
\begin{itemize}
    \item \textit{Insufficient training data}: Although large volumes of RNA and protein sequences are available, experimentally verified interaction data remain relatively scarce, limiting the generalization ability of supervised models.
    \item \textit{Diverse interaction patterns}: RPIs involve multiple types (binding~\cite{xiao2024profiling}, regulation~\cite{nakanishi2021light}, catalysis~\cite{strecker2022rna}, \textit{etc}.), whereas most existing methods treat them as a single homogeneous relation, thereby losing important semantic distinctions.
    \item \textit{Cold-start problem}: When a new RNA or protein appears without known interaction edges, conventional GNNs fail to produce meaningful embeddings, leading to degraded prediction performance.
\end{itemize}

To address these challenges, we propose EGRL, which integrates four key components: 
(1) an implicit meta-path learning module that automatically discovers and aggregates relation-level semantics from multiple edge types to enrich node representations, alleviating data sparsity without requiring additional labels; 
(2) a multi-relational GAT that computes separate attention weights for each edge type (\textit{e.g.}, RNA-RNA similarity, protein-protein similarity, known RPIs, and generated soft edges), thereby preserving interaction heterogeneity; 
(3) a graph generator that predicts probabilistic soft edges between new and existing nodes using only sequence features, trained with a pseudo cold-start auxiliary loss to enable dynamic integration of unseen nodes during inference; and 
(4) an interaction predictor with multi-feature fusion, which captures both linear and multiplicative relationships for final interaction scoring. 
By integrating these components, EGRL achieves competitive performance on benchmark datasets and demonstrates strong generalization in cold-start scenarios.

\begin{figure*}[!t]
    \centering
    \includegraphics[width = \textwidth]{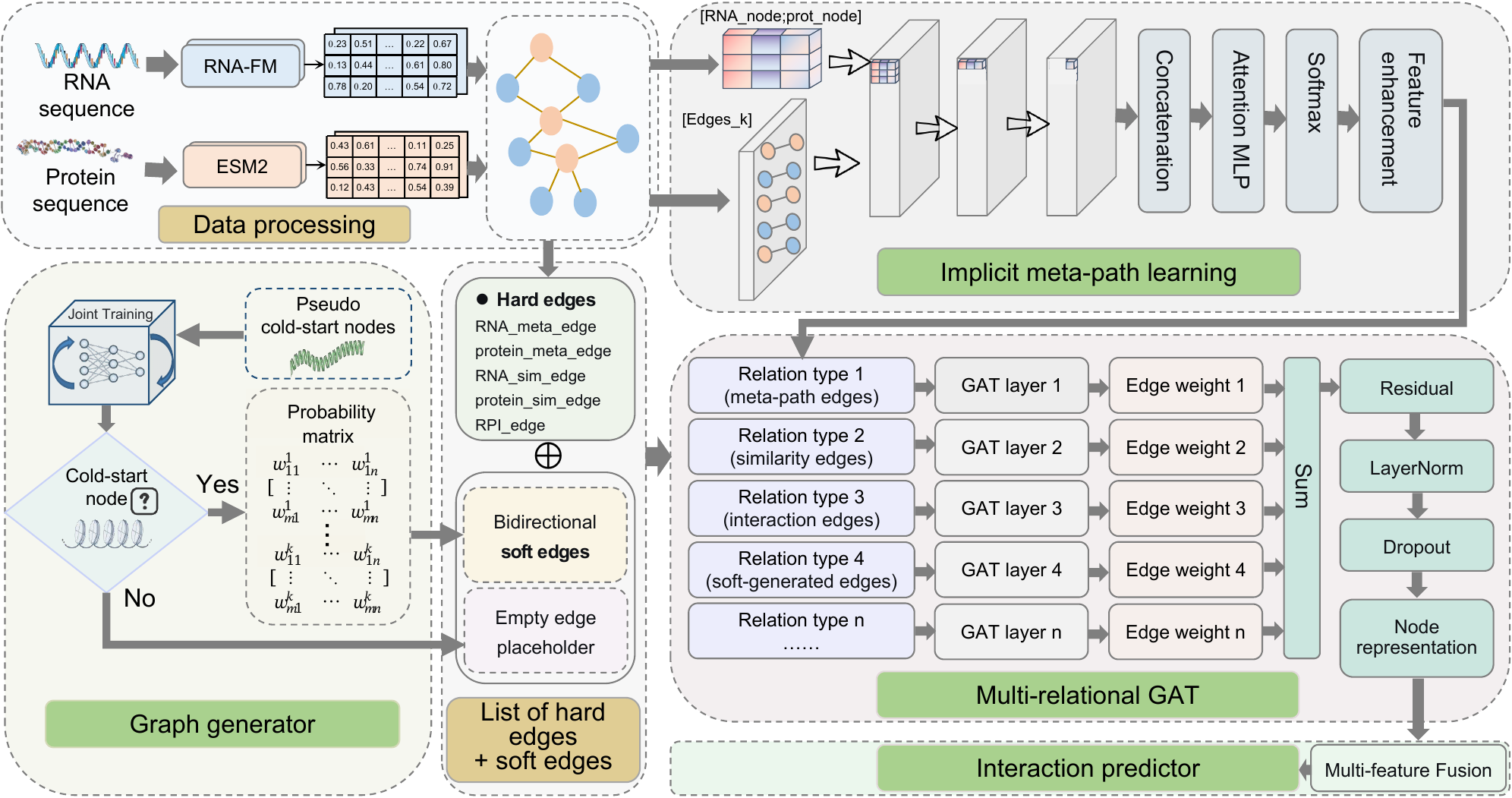}
    \caption{\textbf{Overall architecture of EGRL.} Illustration of the multi-relational graph modeling, implicit meta-path learning, and generator-based enhancement.}
    \label{FIG:architecture}
\end{figure*}

\section{Proposed method}
\label{SEC:Methodology}

We present the architecture of EGRL, as illustrated in Figure~\ref{FIG:architecture}. The framework consists of four key components: (1) \textbf{Implicit meta-path learning}, which automatically learns the importance of different relational paths; (2) \textbf{Graph generator}, which dynamically synthesizes soft edges for cold-start nodes; (3) \textbf{Multi-relational GAT}, which propagates information across multiple edge types; and (4) \textbf{Interaction predictor} with multi-feature fusion, which combines node embeddings for final interaction scoring.

\subsection{Implicit meta-path learning}
\label{Implicit}

Given a heterogeneous graph with node set $\mathcal{V}  =  \mathcal{R} \cup \mathcal{P}$ (RNAs and proteins) and multiple edge types $\mathcal{R}  =  \{ r_1, r_2, \dots, r_K \}$, we aim to capture the semantic importance of each meta-path without manual design. Let $\bm{X} \in \mathbb{R}^{N \times d}$ denote the initial node feature matrix (after linear projection), where $N  =  |\mathcal{R}| + |\mathcal{P}|$ and $d$ is the hidden dimension.

For each relation type $r_k$, we define the set of involved nodes as:
\begin{align}
    \hat{\mathcal{N}}_k  =  \{ i \mid \left(i, j \right) \in \mathcal{E}_k \mathrm{~or~}  \left(j, i \right) \in \mathcal{E}_k \},
\label{formula:nodes_set}
\end{align}
where $\mathcal{E}_k$ denotes the edge set of type $k$. A relation-specific linear transformation $\bm{W}_k \in \mathbb{R}^{d \times d}$ is applied, and the transformed features are averaged to obtain the relation representation:
\begin{align}
    \bm{m}_k  =  \frac{1}{\lvert \hat{\mathcal{N}}_k \rvert} \sum_{i \in \hat{\mathcal{N}}_k} \bm{W}_k \bm{X}[i, :].
\label{formula:relation1}
\end{align}

The importance of each relation is computed via an attention network operating on concatenated representations:
\begin{align}
    \bm{\alpha}  =  \mathrm{Softmax}\left( \mathrm{MLP}\left( \left[ \bm{m}_1 \parallel \dots \parallel \bm{m}_K \right] \right) \right),
    \label{formula:relation2}
\end{align}
where $\parallel$ denotes concatenation and the \textit{Multilayer Perceptron} (MLP) consists of two linear layers with ReLU and dropout. The aggregated meta-path feature is:
\begin{align}
    \bm{h}_{\mathrm{meta}}  =  \sum_{k = 1}^K \bm{\alpha}_k \bm{m}_k.
    \label{formula:relation3}
\end{align}

Finally, $\bm{h}_{\mathrm{meta}}$ is broadcast to all nodes and added to the original features via a residual connection:
\begin{align}
    \bm{X}'  =  \bm{X} + \mathrm{Dropout} \left(\bm{h}_{\mathrm{meta}} \right).
\label{formula:relation4}
\end{align}
This enriches node representations with relation-level semantics while preserving node-specific information.

\subsection{Graph generator for cold-start nodes}

To handle cold-start scenarios where new RNAs or proteins appear without known interactions, we introduce a graph generator that predicts probabilistic edges between new and existing nodes. The generator is a pairwise MLP that takes node features as input and outputs interaction probabilities.

Let $\bm{X}_{\mathrm{new}} \in \mathbb{R}^{M \times d}$ denote the embeddings of $M$ cold-start nodes (after the same feature projection and implicit meta-path enhancement as in \cref{Implicit}), and let $\bm{X}_{\mathrm{old}} \in \mathbb{R}^{N_{\mathrm{old}} \times d}$ denote the embeddings of existing nodes. For each pair $(i,j)$, the interaction probability is computed as:
\begin{align}
    p_{ij}  =  \sigma\left( \mathrm{MLP}_{\mathrm{gen}}\left( \left[ \bm{X}_{\mathrm{new}}[i, :] \parallel \bm{X}_{\mathrm{old}}[j, :] \right] \right) \right),
    \label{eq:Generator1}
\end{align}
where $\parallel$ denotes concatenation and $\sigma$ is the sigmoid function. $\mathrm{MLP}_{\mathrm{gen}}$ consists of two hidden layers with ReLU activation followed by a linear output layer. The output is a probability matrix $\bm{P} \in [0,1]^{M \times N_{\mathrm{old}}}$.

To incorporate these predicted interactions into the graph, we construct bidirectional soft edges with weights equal to the predicted probabilities:
\begin{align}
    \bm{I}_{\mathrm{edge\_index}}  =  \begin{bmatrix} \bm{c} \otimes \bm{1}_{N_{\mathrm{old}}} & \bm{1}_M \otimes \bm{t} \\ \bm{1}_M \otimes \bm{t} & \bm{c} \otimes \bm{1}_{N_{\mathrm{old}}} \end{bmatrix}, \quad
    \bm{W}_{\mathrm{edge\_weight}}  =  \begin{bmatrix} \mathrm{flatten} \left(\bm{P} \right) \\ \mathrm{flatten} \left(\bm{P} \right) \end{bmatrix},
    \label{eq:Generator2}
\end{align}
where $\bm{c}$ and $\bm{t}$ denote the indices of cold-start and existing nodes, respectively, and $\otimes$ denotes a Kronecker product-like expansion that enumerates all pairwise connections. These soft edges are appended to the existing edge list and treated as an additional relation type in subsequent GAT layers.

\textbf{Training Strategy for the Graph Generator}: To ensure that the generator produces meaningful soft edges, we jointly train it with the main model using a self-supervised auxiliary loss. During each training epoch, a subset of RNA and protein nodes is randomly selected as pseudo cold-start nodes. For each such node, the generator predicts interaction probabilities with all candidate nodes, and a binary cross-entropy loss is computed against the ground-truth interaction matrix:
\begin{align}
    \mathcal{L}_{\mathrm{gen}}  =  \frac{1}{M_{\mathrm{rna}}} \sum_{i \in \mathcal{C}_{\mathrm{rna}}} \mathrm{BCE}\left(\bm{P}_{i,\cdot}, \bm{Y}_{i,\cdot}\right) + \frac{1}{M_{\mathrm{prot}}} \sum_{j \in \mathcal{C}_{\mathrm{prot}}} \mathrm{BCE}\left(\bm{P}_{j,\cdot}, \bm{Y}_{j,\cdot}\right),
    \label{eq:Generator3}
\end{align}
where $\mathcal{C}_{\mathrm{rna}}$ and $\mathcal{C}_{\mathrm{prot}}$ denote the pseudo cold-start RNA and protein sets, respectively, $\bm{P}$ is the predicted probability matrix, and $\bm{Y}$ is the corresponding submatrix of the training interaction matrix. This auxiliary loss is combined with the main prediction loss using a weighting factor $\lambda_{\mathrm{gen}}$ (see \cref{formula:loss2}) and optimized jointly.

\subsection{Multi-relational GAT}

The enhanced node features $\bm{X}'$ (after meta-path learning, \cref{formula:relation4}) are fed into a multi-relational GAT to perform relation-aware message passing. We consider multiple edge types, including original hard edges (RNA-RNA, protein-protein, similarity, and known RNA-protein interactions) as well as soft edges generated for cold-start nodes (during both training and inference).

For each relation type $r$, an independent GAT convolution is applied:
\begin{align}
    \bm{h}_r  =  \mathrm{GATConv}_r\left( \bm{X}', \mathcal{E}_r, \mathrm{edge\_attr}  =  \bm{W}_r \right),
    \label{formula:Relational1} 
\end{align}
where $\mathcal{E}_r$ denotes the edge index of relation $r$, and $\bm{W}_r$ is the corresponding edge weight vector (for soft edges) or omitted for hard edges. Each $\mathrm{GATConv}$ employs $h$ attention heads and outputs features of dimension $d$.

The outputs from all relations are averaged and combined with a residual connection, followed by layer normalization:
\begin{align}
    \bm{X}^{(l+1)}  =  \mathrm{LayerNorm}\left( \bm{X}^{(l)} + \mathrm{Dropout}\left( \frac{1}{\lvert \mathcal{R} \rvert} \sum_{r = 1}^{\lvert \mathcal{R} \rvert} \bm{h}_r \right) \right),
    \label{formula:Relational2}
\end{align}
where $\bm{X}^{(0)}  =  \bm{X}'$ and $\mathcal{R}$ denotes the set of all relation types (including soft edges). This process is repeated for $L$ layers, enabling the model to capture both local and multi-hop dependencies.

\subsection{Interaction predictor with multi-feature fusion}

After the final GAT layer, we obtain embeddings for RNAs and proteins: $\bm{h}_{\mathcal{R}} \in \mathbb{R}^{\lvert \mathcal{R} \rvert \times d}$ and $\bm{h}_{\mathcal{P}} \in \mathbb{R}^{\lvert \mathcal{P} \rvert \times d}$. For a given RNA-protein pair $(r,p)$, we construct a feature vector:
\begin{align}
    \bm{f}_{\mathrm{cross}}  =  \left[ \bm{h}_{\mathcal{R}}[r] \parallel \bm{h}_{\mathcal{P}}[p] \parallel \bm{h}_{\mathcal{R}}[r] \odot \bm{h}_{\mathcal{P}}[p] \parallel \bm{h}_{\mathcal{R}}[r] - \bm{h}_{\mathcal{P}}[p] \right] \in \mathbb{R}^{4d},
    \label{formula:Predictor1}
\end{align}
where $\odot$ denotes element-wise product. This formulation captures both linear and multiplicative relationships.

The feature vector is then passed through a two-layer predictor network with layer normalization and ReLU activation:
\begin{align}
    \hat{y}  =  \sigma\left(\bm{W}_2 \cdot \mathrm{ReLU}\left( \mathrm{LayerNorm}\left( \bm{W}_1 \bm{f}_{\mathrm{cross}} + \bm{b}_1 \right) \right) + \bm{b}_2 \right),
    \label{formula:Predictor2}
\end{align}
where $\sigma$ is the sigmoid function, $\bm{W}_1 \in \mathbb{R}^{d \times 4d}$ and $\bm{W}_2 \in \mathbb{R}^{1 \times d}$ are weight matrices, and $\bm{b}_1$, $\bm{b}_2$ are bias vectors. The output $\hat{y} \in [0,1]$ denotes the predicted interaction probability.

\subsection{Joint training objective}

EGRL is trained by minimizing a weighted binary cross-entropy loss for the main prediction task, combined with a generator auxiliary loss. Let $y_i \in [0,1]$ denote the ground-truth label for the $i$-th RNA-protein pair. The main loss is defined as:
\begin{align}
    \mathcal{L}_{\mathrm{main}}  =  -\frac{1}{N} \sum_{i = 1}^N \left[ w_p \cdot y_i \log \hat{y}_i + \left(1 - y_i \right) \log \left(1 - \hat{y}_i \right) \right],
    \label{formula:loss1}
\end{align}
where $w_p$ is the weight assigned to positive samples. In practice, $w_p  =  \beta \cdot \frac{\mathrm{neg}}{\mathrm{pos}}$, where $\beta$ further emphasizes positive interactions.

The total loss is given by:
\begin{align}
    \mathcal{L}_{\mathrm{total}}  =  \mathcal{L}_{\mathrm{main}} + \lambda_{\mathrm{gen}} \mathcal{L}_{\mathrm{gen}},
    \label{formula:loss2}
\end{align}
where $\lambda_{\mathrm{gen}}$ is a hyperparameter controlling the contribution of the generator loss $\mathcal{L}_{\mathrm{gen}}$ (see \cref{eq:Generator3}). This joint optimization encourages the generator to learn meaningful soft edges from node features, enabling effective interaction prediction for unseen nodes during inference.



\section{Experimental setup}

\label{SEC:Experiment}
We present the experimental setup, including research questions (RQs), datasets, comparison methods, and evaluation metrics.

\subsection{Research questions}
\label{SEC:RQ}

To evaluate the proposed approach, we investigate the following research questions (RQs):

\textit{\textbf{1) RQ1:} How do the core architectural components of EGRL affect performance?} 

To quantify the contribution of each module, we conduct an ablation study with five configurations:
\begin{itemize}
    \item Full: All modules enabled.
    \item -A (w/o ImplicitMetaPath): Removes the implicit meta-path learner, directly feeding node features into the GAT.
    \item -B (w/o MultiRelationalGAT): Replaces the multi-relational GAT with a single-relation GAT that merges all edge types, disabling relation-aware attention.
    \item -C (w/o GraphGenerator): Disables the graph generator; no soft edges are introduced for cold-start nodes.
    \item -D (w/o Multi-featureFusion): Removes multi-feature fusion (Hadamard product and absolute difference), using only concatenated embeddings.
\end{itemize}
This design isolates the contribution of each component while controlling for model capacity.

\textit{\textbf{2) RQ2:} How do key hyperparameters influence the performance of EGRL?} 

We evaluate the robustness of the model with respect to its key hyperparameters.

\textit{Experiment 1: Number of \textit{k}-nearest neighbours (kNN).}  
We vary $k \in \{3,5,10,15,20\}$ to examine how the number of neighbors in the similarity graph affects performance under the normal scenario (without generator-based soft edges). Larger $k$ increases connectivity but may introduce noise.

\textit{Experiment 2: Generator-related hyperparameters.}  
Under the cold-start scenario, all test nodes are treated as unseen, and the generator produces soft edges. During training, a pseudo cold-start strategy is applied. We vary:
\begin{itemize}
    \item Generator loss weight $\lambda_{\mathrm{gen}} \in \{0.02, 0.05, 0.1, 0.2\}$, balancing generator learning and main task performance.
    \item Pseudo cold-start node ratio $\in \{0.01, 0.05, 0.1, 0.15\}$, controlling the number of simulated unseen nodes.
\end{itemize}
  
\textit{\textbf{3) RQ3:} How does EGRL compare with state-of-the-art methods?} 

We compare EGRL with representative RPIP methods on benchmark datasets, including RLF-LPI~\cite{song2022rlf}, RPI-SAN~\cite{yi2018deep}, RPITER~\cite{peng2019rpiter}, IPMiner~\cite{pan2016ipminer}, and NPI-GNN~\cite{shen2021npi} (see \cref{other-methods}).
    
\textit{\textbf{4) RQ4:} Can EGRL generalize to completely unseen RNA and protein families?} 

To assess generalization under cold-start conditions, we conduct two evaluations:

\textit{Experiment 1: Molecule hold-out.}  
Selected molecules are entirely removed from training, and their interactions are predicted at test time using generator-produced soft edges.

\textit{Experiment 2: Sequence-cluster-based partitioning.}  
To prevent data leakage, RNA and protein sequences are clustered using CD-HIT~\citep{li2006cd} with identity thresholds of 80\% (RNA) and 40\% (protein). Entire clusters are held out as test data, ensuring no high-similarity overlap with training samples. This setting evaluates generalization to novel sequence families under a strict cold-start scenario.

\begin{table}[!t]
    \footnotesize
    \centering
    \caption{\textbf{Statistics of the four benchmark RNA-protein interaction (RPI) datasets.}}
    \label{tab:datasets}
    \setlength{\tabcolsep}{12pt}
    \begin{tabular}{l|cccc}
    \toprule[1.1pt]
    {Dataset} & {\# RNAs} & {\# Proteins} & {\# Interactions} & {\# Non-interactions} \\
    \midrule
    RPI369~\cite{muppirala2011predicting} & 332 & 338 & 369 & 0 \\
    RPI1807~\cite{suresh2015rpi} & 1,078 & 3,131 & 1,807 & 1,436 \\
    RPI2241~\cite{muppirala2011predicting} & 841 & 2,042 & 2,241 & 0 \\
    NPInter2~\cite{yuan2014npinter} & 4,636 & 449 & 10,412 & 0 \\
    \bottomrule[1.1pt]
    \end{tabular}
\end{table}

\subsection{Datasets}

We utilize four benchmark datasets from prior studies~\cite{peng2019rpiter}, as summarized in \cref{tab:datasets}, to evaluate our method. The datasets RPI369, RPI1807, and RPI2241 are derived from RNA-protein complexes collected from the RNA-Protein Interaction Database (PRIDB)~\cite{lewis2010pridb} and the Protein Data Bank (PDB)~\cite{berman2000protein}. These datasets are constructed based on the minimum atomic distance criterion, where an RNA-protein pair is considered interacting if the distance between any RNA atom and protein atom falls below a predefined threshold. RPI369 is a subset of RPI2241~\cite{muppirala2011predicting}, excluding interactions involving ribosomal proteins or ribosomal RNA. The RPI1807 dataset is obtained by parsing data from the Nucleic Acid Database (NDB)~\cite{berman2003nucleic} and PRIDB, which provide RNA-protein complex structures and atomic interaction interfaces.

In contrast, the NPInter2 dataset consists of experimentally validated ncRNA-protein interactions, particularly involving lncRNAs, collected from the NPInter2 database~\cite{yuan2014npinter}, rather than being defined by structural distance criteria.

Since RPI369, RPI2241, and NPInter2 do not provide negative samples, we follow~\cite{peng2019rpiter} to generate an equal number of non-interacting pairs by randomly combining RNAs and proteins from positive samples. To reduce false negatives, pairs are discarded if they are highly similar to known interactions; specifically, a generated pair $(R_1, P_1)$ is removed if there exists $(R_2, P_2)$ such that the sequence identity between $R_1$ and $R_2$ exceeds 80\% and that between $P_1$ and $P_2$ exceeds 40\%.

\subsection{Previous RPIP methods under comparison}
\label{other-methods}

We evaluate EGRL on the four datasets by comparing it with representative methods in RPIP. The following approaches are considered:

\begin{itemize}
    \item \textbf{RLF-LPI}~\cite{song2022rlf}: An ensemble framework that integrates a long short-term memory autoencoder with an attention mechanism to extract latent representations, combined with a fuzzy decision-making strategy to model lncRNA-protein relationships.
    
    \item \textbf{RPI-SAN}~\cite{yi2018deep}: A sequence-based method that combines deep learning techniques with a random forest classifier, leveraging RNA sequences and protein evolutionary information for prediction.
    
    \item \textbf{RPITER}~\cite{peng2019rpiter}: A multi-level deep learning framework that integrates convolutional neural networks and stacked autoencoders, exploiting both sequence and structural information of RNA and proteins.
    
    \item \textbf{IPMiner}~\cite{pan2016ipminer}: A stacked autoencoder-based approach that learns latent interaction patterns from RNA and protein sequences, followed by a random forest classifier within an ensemble framework.
    
    \item \textbf{NPI-GNN}~\cite{shen2021npi}: An end-to-end graph neural network-based method for RPIP, which models interactions using both network structure and sequence information.
\end{itemize}

\subsection{Evaluation metrics}

We adopt the commonly used five-fold cross-validation to evaluate model performance. Multiple metrics are employed, including \textit{Accuracy} (ACC), \textit{Precision} (PREC), \textit{Recall} (REC), \textit{F1-score} (F1), and \textit{Matthews Correlation Coefficient} (MCC). In addition, threshold-independent metrics, namely the \textit{Area Under the Receiver Operating Characteristic Curve} (AUROC) and the \textit{Area Under the Precision-Recall Curve} (AUPR), are also reported.

The metrics are defined as follows:
\begin{align}
    \mathrm{ACC} & =  \frac{\mathrm{TP} + \mathrm{TN}}{\mathrm{TP} + \mathrm{FP} + \mathrm{TN} + \mathrm{FN}}, \label{formula:Acc} \\
    \mathrm{PREC} & =  \frac{\mathrm{TP}}{\mathrm{TP} + \mathrm{FP}}, \label{formula:pre} \\
    \mathrm{REC} & =  \frac{\mathrm{TP}}{\mathrm{TP} + \mathrm{FN}}, \label{formula:Se} \\
    \mathrm{MCC} & =  \frac{\mathrm{TP} \times \mathrm{TN} - \mathrm{FP} \times \mathrm{FN}}{\sqrt{\left(\mathrm{TP} + \mathrm{FP} \right) \left(\mathrm{TP} + \mathrm{FN} \right) \left(\mathrm{TN} + \mathrm{FP} \right) \left(\mathrm{TN} + \mathrm{FN} \right)}}, \label{formula:mcc} \\
    \mathrm{F1} & =  2 \cdot \frac{\mathrm{PREC} \times \mathrm{REC}}{\mathrm{PREC} + \mathrm{REC}}, \label{formula:F1} \\
    \mathrm{AUROC} & =  \int_{0}^{1} \mathrm{TPR} \left(\mathrm{FPR} \right)  d \left(\mathrm{FPR} \right), \label{formula:AUROC} \\
    \mathrm{AUPR} & =  \int_{0}^{1} \mathrm{PREC}(\mathrm{REC})  d(\mathrm{REC}), \label{formula:AUPR}
\end{align}
where $\mathrm{TP}$, $\mathrm{FP}$, $\mathrm{TN}$, and $\mathrm{FN}$ denote the numbers of true positives, false positives, true negatives, and false negatives, respectively. $\mathrm{TPR}$ denotes the true positive rate (recall), and $\mathrm{FPR}$ is defined as:
\begin{align}
    \mathrm{FPR}  =  \frac{\mathrm{FP}}{\mathrm{FP} + \mathrm{TN}}. \label{formula:FPR}
\end{align}

\subsection{Hardware and software environment}

EGRL is implemented in Python using PyTorch 2.6.0+cu118 and PyTorch Geometric 2.6.1. The input features for RNA and protein nodes are obtained from pre-trained language models: RNA-FM provides 256-dimensional embeddings for RNAs, and ESM2-t33-650M produces 1280-dimensional embeddings for proteins. These features are projected into a unified 128-dimensional hidden space via linear layers.

The hidden dimension of both the meta-path processor and the relation-aware network is set to 128. The model is trained using the AdamW optimizer with an initial learning rate of $1\times10^{-4}$ and a cosine annealing scheduler. Training is performed for 300 epochs with a batch size of 256. A weighted binary cross-entropy loss is employed, where the positive sample weight is dynamically set to $10 \cdot \frac{\mathrm{neg}}{\mathrm{pos}}$ to address class imbalance. Gradient clipping is applied with a maximum norm of 1.0, and dropout is set to 0.2. All experiments are conducted on a single NVIDIA V100 (32GB) GPU using five-fold cross-validation.

 
\begin{figure*}
    \centering
    \subfigure[On RPI369 dataset.\label{RPI369 dataset}]{
    \includegraphics[width = 0.7\textwidth]{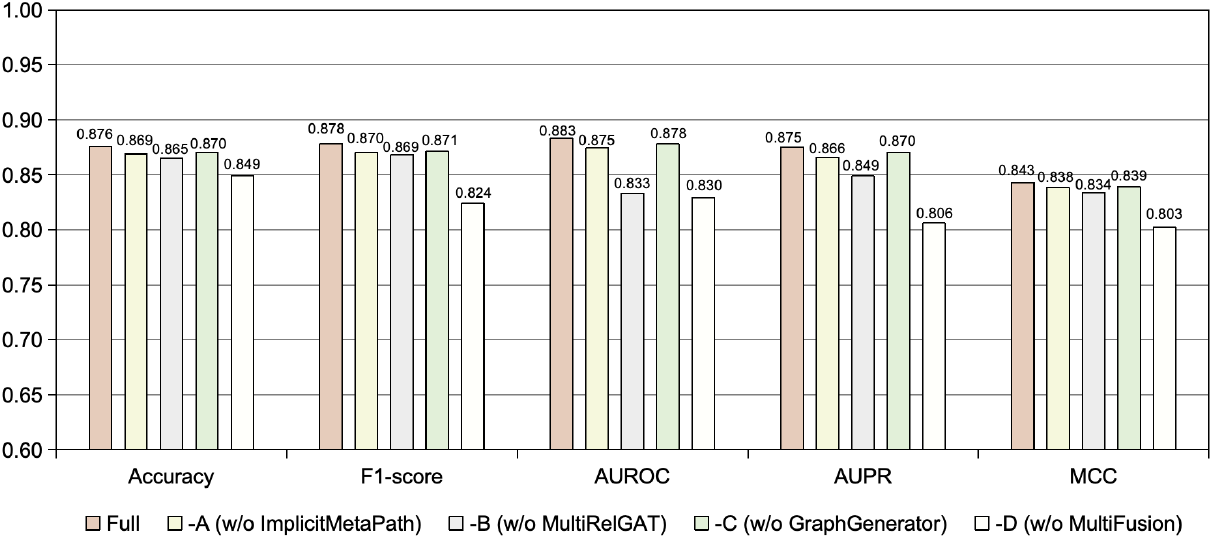}}
    \subfigure[On RPI1807 dataset.\label{RPI1807 dataset}]{
    \includegraphics[width = 0.7\textwidth]{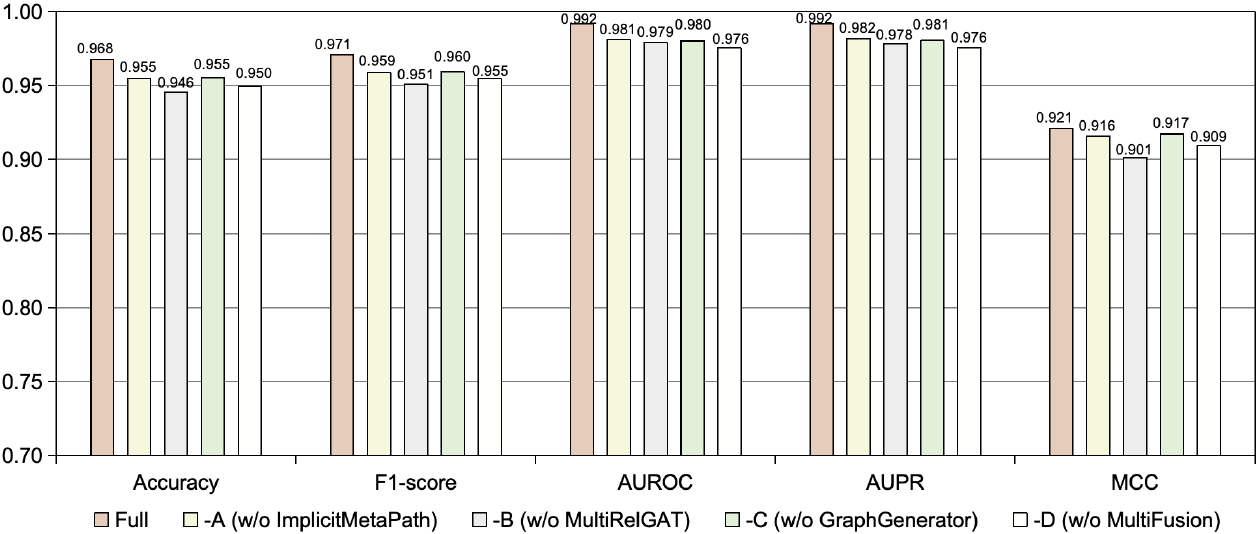}}
    \subfigure[On RPI2241 dataset.\label{RPI2241 dataset}]{
    \includegraphics[width = 0.7\textwidth]{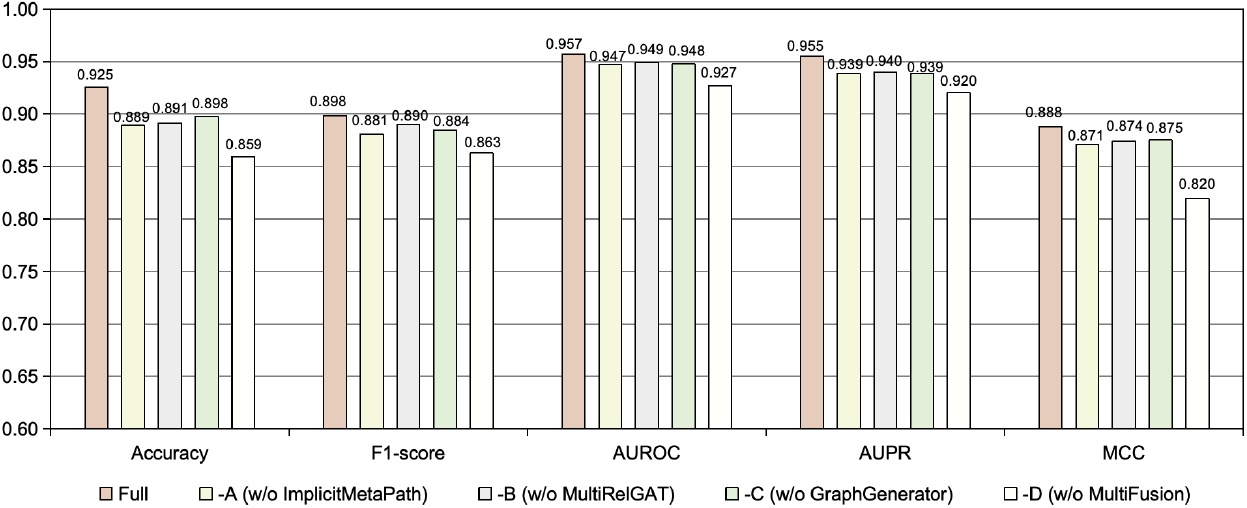}}
    \subfigure[On NPInter2 dataset.\label{NPInter2 dataset}]{
    \includegraphics[width = 0.7\textwidth]{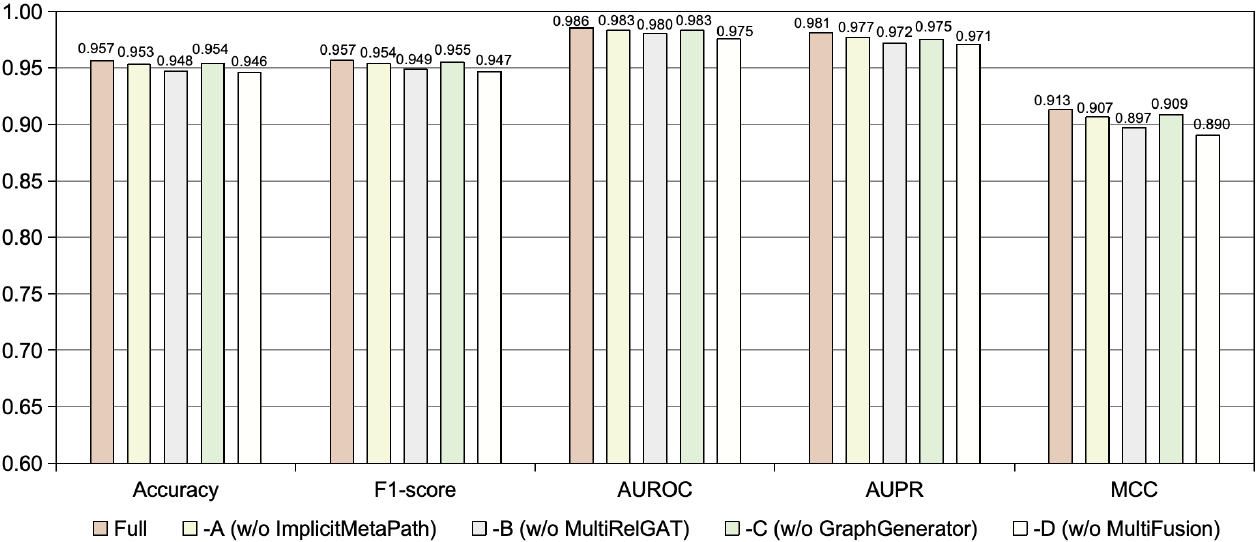}}
    \caption{\textbf{Ablation study on four datasets.} Performance comparison of different EGRL variants under component removal.}
    \label{FIG:Ablation}
\end{figure*}

\section{Experimental results}
\label{SEC:Results}
In this section, we mainly provide the experimental results to answer the above RQs.

\subsection{Answer to RQ1: Ablation study}

The ablation results on the four datasets are shown in Figure~\ref{FIG:Ablation}. Overall, the core components of EGRL contribute to the final performance to different extents. Removing the implicit meta-path module (-A) decreases the Accuracy on RPI2241 from 0.925 to 0.889. For the graph generator (-C), its removal leads to only slight performance degradation, e.g., on RPI369 (Accuracy: 0.876 $\rightarrow$ 0.870, F1: 0.878 $\rightarrow$ 0.871) and RPI1807 (Accuracy: 0.968 $\rightarrow$ 0.955), suggesting limited benefit under the five-fold cross-validation setting, which does not fully reflect strict cold-start scenarios.

In contrast, removing the multi-feature fusion module (-D) results in substantial degradation across all metrics. On RPI369, Accuracy drops from 0.876 to 0.849 and AUPR from 0.875 to 0.806; on RPI2241, Accuracy decreases from 0.925 to 0.859 and MCC from 0.888 to 0.820. This indicates that simple concatenation is insufficient to capture effective interactions, while the Hadamard product and absolute difference play a key role in modeling feature interactions. The multi-relational GAT module (-B) is also critical: removing it reduces AUROC on RPI369 from 0.883 to 0.833 and Accuracy on RPI1807 from 0.968 to 0.946, highlighting the importance of relation-aware modeling in heterogeneous graphs.

Notably, all ablated variants contain fewer parameters than the full model. Overall, the complete EGRL consistently achieves the best performance, indicating that its effectiveness arises from the coordinated design of multiple modules. The multi-feature fusion enhances discriminative capability, while the multi-relational GAT captures structural information. The implicit meta-path module and graph generator further improve generalization from semantic and data augmentation perspectives, respectively.

\begin{tcolorbox}[colback = white,sharp corners,boxrule = 1pt]
    \textit{\textbf{Summary of answers to RQ1:}}
    All ablated variants underperform the full model, with varying degrees of degradation, confirming that EGRL benefits from the synergistic design of its components rather than parameter scaling alone.
\end{tcolorbox}

\subsection{Answer to RQ2: Impacts of hyperparameter tuning}

Following \cref{SEC:RQ}, we conduct systematic analyses on three key hyperparameters: the number of kNN neighbors ($k$), the generator loss weight ($\lambda_{\mathrm{gen}}$), and the pseudo cold-start node ratio (\textit{cold-ratio}).

\begin{figure*}[!t]
    \centering
    \includegraphics[width = \textwidth]{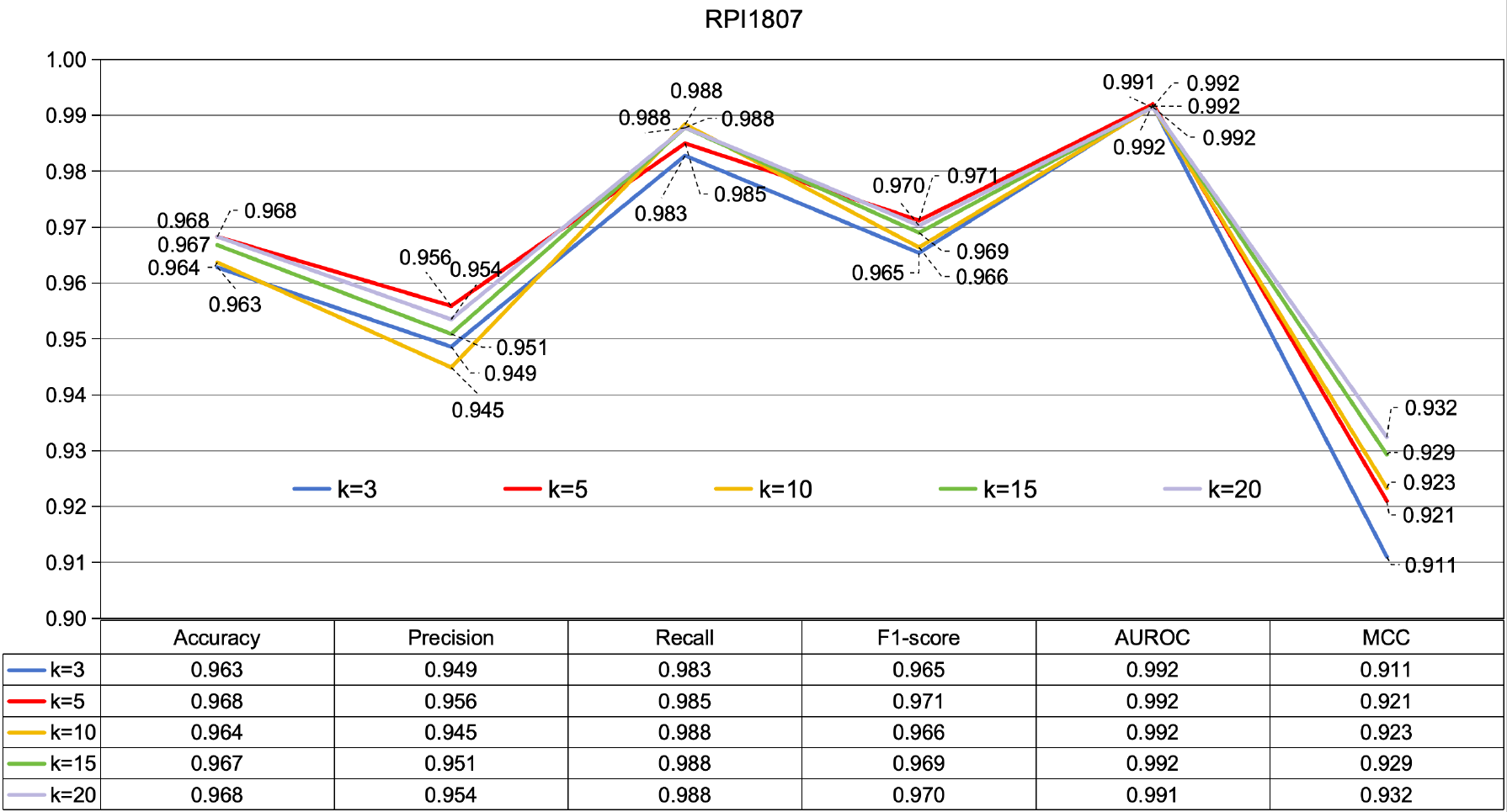}
    \caption{\textbf{Effect of kNN neighbors.} Impact of the number of neighbors $k$ on model performance under the standard evaluation setting.}
    \label{FIG:knn}
\end{figure*}

We first examine the effect of $k$ on model performance. As shown in Figure~\ref{FIG:knn}, performance on the RPI1807 dataset remains stable across different values of $k$. While AUROC stays consistently high as $k$ increases from 3 to 20, Recall improves from 0.983 to 0.988, and both F1-score and MCC show a gradual increase (with MCC peaking at 0.932 for $k = 20$). Precision slightly decreases at $k = 10$ (0.945), indicating that enlarging the neighborhood captures more positives but may introduce noise, reflecting a precision-recall trade-off. Overall, $k = 5$ provides a balanced configuration in terms of Accuracy (0.968) and Precision (0.956), whereas $k = 20$ favors higher Recall and MCC. These results indicate that the model is robust to the choice of neighborhood size under the standard evaluation setting.

\begin{figure*}[!t]
    \centering
    \subfigure[AUROC on RPI1807.\label{1807}]{
    \includegraphics[width = 0.485\textwidth]{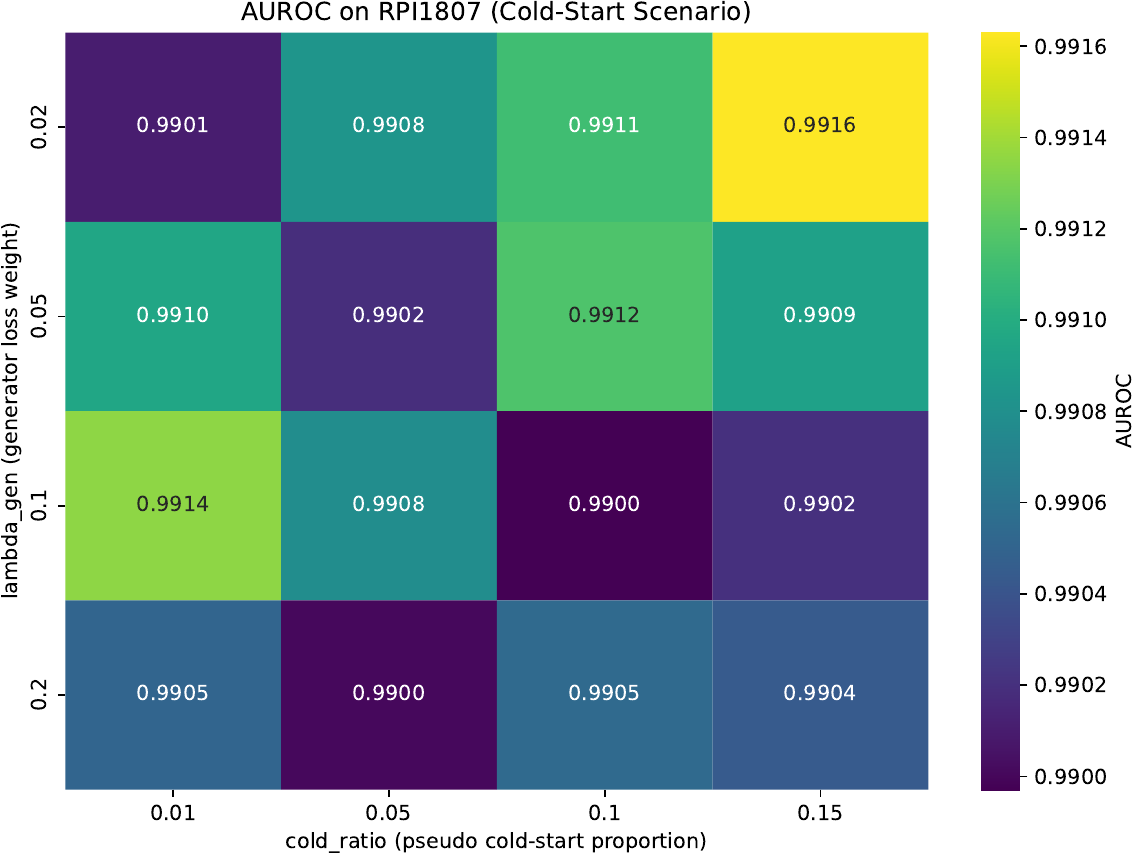}}
    \subfigure[AUROC on RPI2241.\label{2241}]{
    \includegraphics[width = 0.485\textwidth]{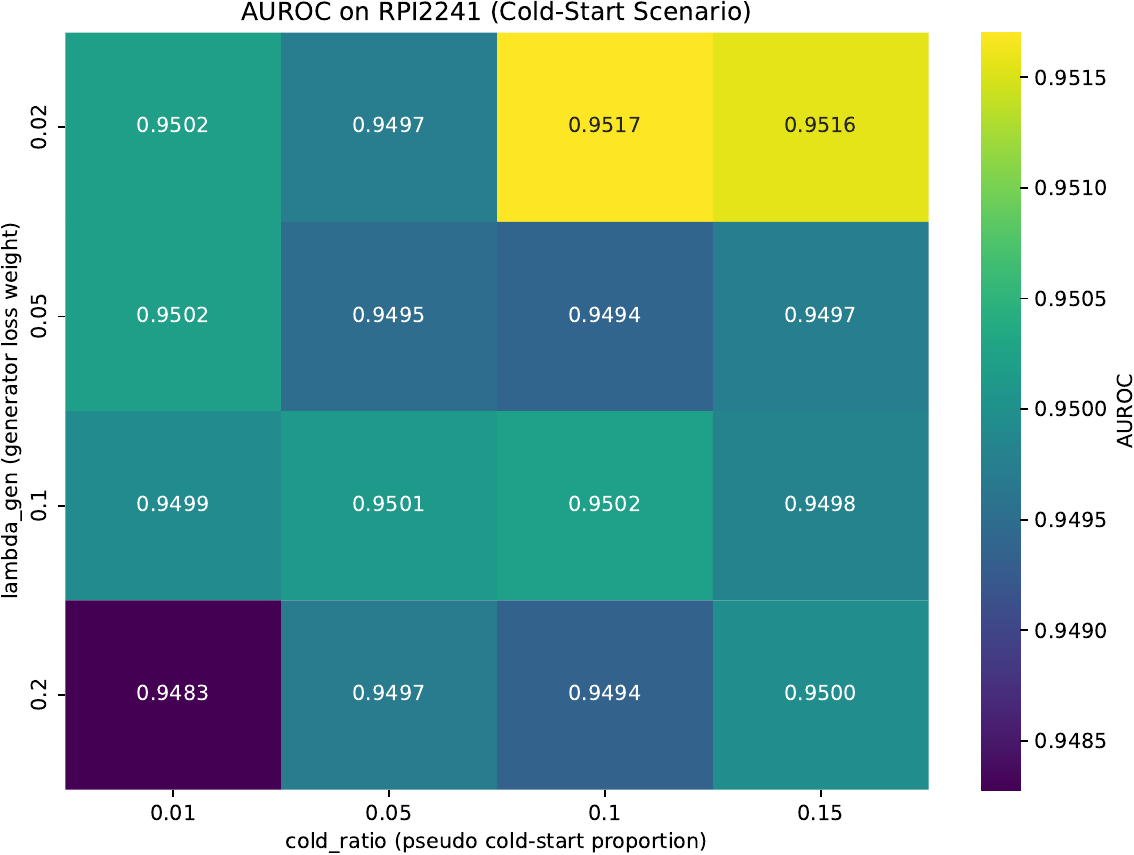}}
    \caption{\textbf{Effect of generator hyperparameters under cold-start scenarios.} Sensitivity of AUROC to $\lambda_{\mathrm{gen}}$ and cold-start ratio.}
    \label{FIG:parameters}
\end{figure*}

In cold-start scenarios, where test nodes are absent from the training graph, the soft-edge completion mechanism of the graph generator becomes critical. We therefore further evaluate the impact of $\lambda_{\mathrm{gen}}$ and \textit{cold-ratio}. As shown in Figure~\ref{FIG:parameters}, on both RPI1807 and RPI2241, varying $\lambda_{\mathrm{gen}}$ within [0.02, 0.05, 0.1, 0.2] and \textit{cold-ratio} within [0.01, 0.05, 0.1, 0.15] leads to only minor AUROC fluctuations (within 0.0016 and 0.0022, respectively). Although the absolute AUROC on RPI2241 (approximately 0.95) is lower than that on RPI1807 (approximately 0.99), both datasets consistently show low sensitivity to these hyperparameters. This indicates that the generator module operates stably across a wide configuration range and effectively supports cold-start prediction.

In summary, EGRL demonstrates strong robustness across both normal and cold-start scenarios. Its performance remains stable with respect to $k$, adapting well to different local graph structures, and shows minimal sensitivity to generator-related hyperparameters ($\lambda_{\mathrm{gen}}$ and \textit{cold-ratio}). These observations suggest that the model maintains reliable performance over a broad hyperparameter space, reducing the need for extensive tuning.

\begin{tcolorbox}[colback = white,sharp corners,boxrule = 1pt]
    \textit{\textbf{Summary of answers to RQ2:}}
    EGRL exhibits strong robustness to variations in the number of neighbors, the generator loss weight, and the pseudo cold-start node ratio, maintaining stable AUROC performance across settings.
\end{tcolorbox}

\begin{table}[!t]
    \centering
    \footnotesize
    \caption{\textbf{Performance comparison on four datasets under five-fold cross-validation.} The best and second-best results are highlighted in \colorbox{lightgreen}{lightgreen} and \colorbox{lightred}{lightred}, respectively.}
    \label{Tab:other_models}
    \setlength{\tabcolsep}{10pt}
    \begin{tabular}{c|l|ccccc}
    \toprule[1.1pt]
    {Dataset} & {Method} & {ACC} & {PREC} & {REC} & {MCC} & {AUROC} \\
    \midrule
    \multirow{6}{*}{RPI369} 
    & RLF-LPI~\cite{song2022rlf} & \colorbox{lightred}{0.844} & \colorbox{lightred}{0.819} & \colorbox{lightgreen}{0.882} & \colorbox{lightred}{0.610} & \colorbox{lightgreen}{0.905} \\ 
    & RPI-SAN~\cite{yi2018deep} & - & - & - & - & - \\
    & RPITER~\cite{peng2019rpiter} & 0.728 & 0.701 & 0.797 & 0.461 & 0.821 \\ 
    & IPMiner~\cite{pan2016ipminer} & 0.752 & 0.713 & 0.735 & 0.507 & 0.773 \\ 
    & NPI-GNN~\cite{shen2021npi} & 0.602 & 0.600 & 0.615 & 0.212 & - \\  
    & {EGRL (Ours)} & \colorbox{lightgreen}{0.876} & \colorbox{lightgreen}{0.878} & \colorbox{lightred}{0.875} & \colorbox{lightgreen}{0.843} & \colorbox{lightred}{0.883} \\  
    \midrule

    \multirow{6}{*}{RPI1807} 
    & RLF-LPI~\cite{song2022rlf} & \colorbox{lightred}{0.973} & \colorbox{lightred}{0.968} & 0.984 & 0.927 & 0.987 \\ 
    & RPI-SAN~\cite{yi2018deep} & 0.961 & 0.914 & 0.936 & 0.924 & \colorbox{lightgreen}{0.999} \\ 
    & RPITER~\cite{peng2019rpiter} & 0.968 & 0.959 & \colorbox{lightred}{0.986} & \colorbox{lightred}{0.936} & 0.990 \\ 
    & IPMiner~\cite{pan2016ipminer} & \colorbox{lightgreen}{0.986} & \colorbox{lightgreen}{0.978} & 0.982 & \colorbox{lightgreen}{0.972} & \colorbox{lightred}{0.998} \\ 
    & NPI-GNN~\cite{shen2021npi} & - & - & - & - & - \\ 
    & {EGRL (Ours)} & 0.968 & 0.962 & \colorbox{lightgreen}{0.993} & 0.921 & 0.992 \\ 
    \midrule

    \multirow{6}{*}{RPI2241} 
    & RLF-LPI~\cite{song2022rlf} & 0.842 & \colorbox{lightred}{0.874} & 0.800 & 0.549 & 0.924 \\ 
    & RPI-SAN~\cite{yi2018deep} & \colorbox{lightred}{0.907} & 0.840 & 0.861 & \colorbox{lightred}{0.822} & \colorbox{lightgreen}{0.962} \\ 
    & RPITER~\cite{peng2019rpiter} & 0.890 & 0.871 & \colorbox{lightred}{0.917} & 0.781 & \colorbox{lightred}{0.957} \\ 
    & IPMiner~\cite{pan2016ipminer} & 0.824 & 0.836 & 0.833 & 0.650 & 0.906 \\ 
    & NPI-GNN~\cite{shen2021npi} & 0.626 & 0.672 & 0.498 & 0.270 & - \\ 
    & {EGRL (Ours)} & \colorbox{lightgreen}{0.925} & \colorbox{lightgreen}{0.877} & \colorbox{lightgreen}{0.933} & \colorbox{lightgreen}{0.888} & \colorbox{lightred}{0.957} \\  
    \midrule

    \multirow{6}{*}{NPInter2} 
    & RLF-LPI~\cite{song2022rlf} & - & - & - & - & - \\ 
    & RPI-SAN~\cite{yi2018deep} & \colorbox{lightgreen}{0.986} & - & - & - & - \\ 
    & RPITER~\cite{peng2019rpiter} & 0.955 & \colorbox{lightred}{0.939} & \colorbox{lightred}{0.973} & \colorbox{lightred}{0.910} & 0.985 \\ 
    & IPMiner~\cite{pan2016ipminer} & 0.952 & \colorbox{lightgreen}{0.945} & 0.946 & 0.904 & \colorbox{lightgreen}{0.995} \\ 
    & NPI-GNN~\cite{shen2021npi} & 0.933 & 0.915 & 0.956 & 0.868 & 0.970 \\  
    & {EGRL (Ours)} & \colorbox{lightred}{0.957} & \colorbox{lightgreen}{0.945} & \colorbox{lightgreen}{0.977} & \colorbox{lightgreen}{0.913} & \colorbox{lightred}{0.986} \\  
    \bottomrule[1.1pt]
    \end{tabular}
\end{table}

\subsection{Answer to RQ3: EGRL \textit{vs.} State-of-the-art methods}

We evaluate EGRL using Accuracy, Precision, Recall, MCC, and AUROC, with detailed results reported in \cref{Tab:other_models}. Based on comparisons with representative methods, the main observations are as follows.

EGRL achieves competitive performance across all four benchmark datasets. On RPI369, it attains the highest Accuracy, Precision, and MCC, indicating a clear advantage and strong capability in learning from limited data. On RPI1807, EGRL achieves the highest Recall (0.993), while other metrics remain close to the best-performing methods; for instance, its Precision (0.962) and AUROC (0.992) are slightly below those of IPMiner~\cite{pan2016ipminer} (0.978) and RPI-SAN~\cite{yi2018deep} (0.999), respectively, yet remain well balanced overall. On RPI2241, EGRL obtains the highest Accuracy (0.925) and MCC (0.888), with consistently strong performance across other metrics, demonstrating robustness on larger datasets. On NPInter2, EGRL achieves the highest Recall (0.977) and maintains competitive AUROC (0.986) and MCC (0.913), further validating its effectiveness across diverse data distributions.

\begin{tcolorbox}[colback = white,sharp corners,boxrule = 1pt]
    \textit{\textbf{Summary of answers to RQ3:}}
    EGRL consistently delivers competitive performance across datasets of different scales and characteristics, demonstrating robust and well-balanced predictive capability.
\end{tcolorbox}

\begin{figure*}[!t]
    \centering
    \subfigure[AUROC of molecule hold-out validation.\label{AUC_NPInter}]{
    \includegraphics[width = 0.485\textwidth]{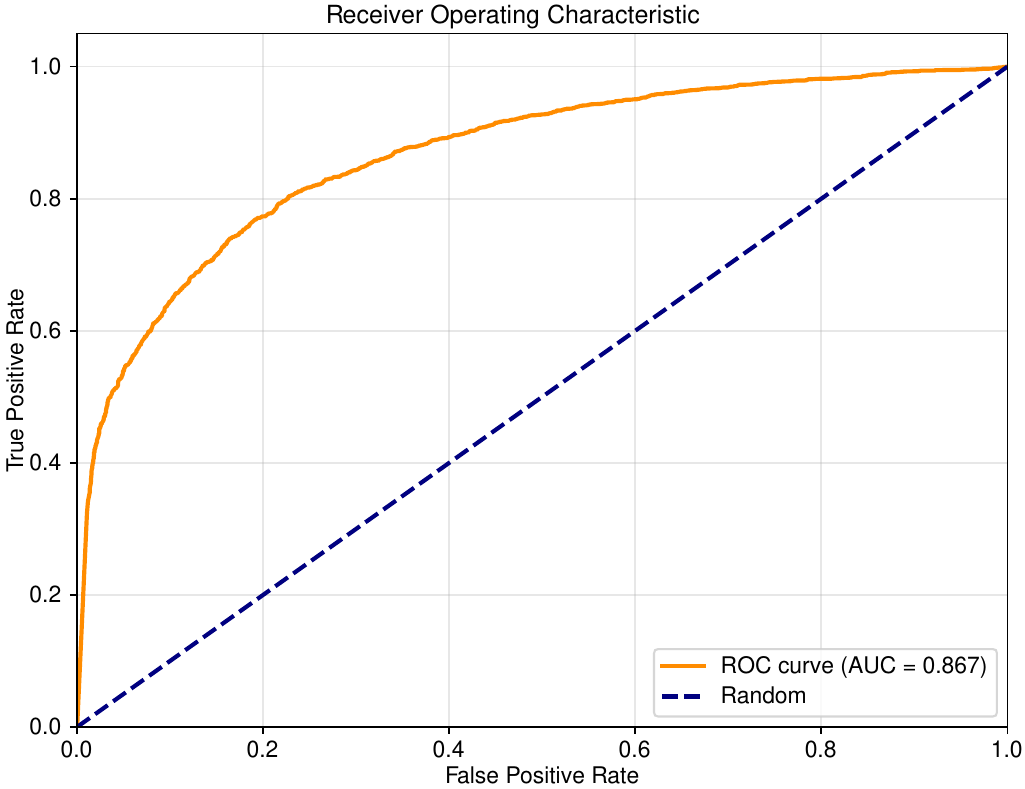}}
    \subfigure[AUPR molecule hold-out validation.\label{AUPR_NPInter}]{
    \includegraphics[width = 0.485\textwidth]{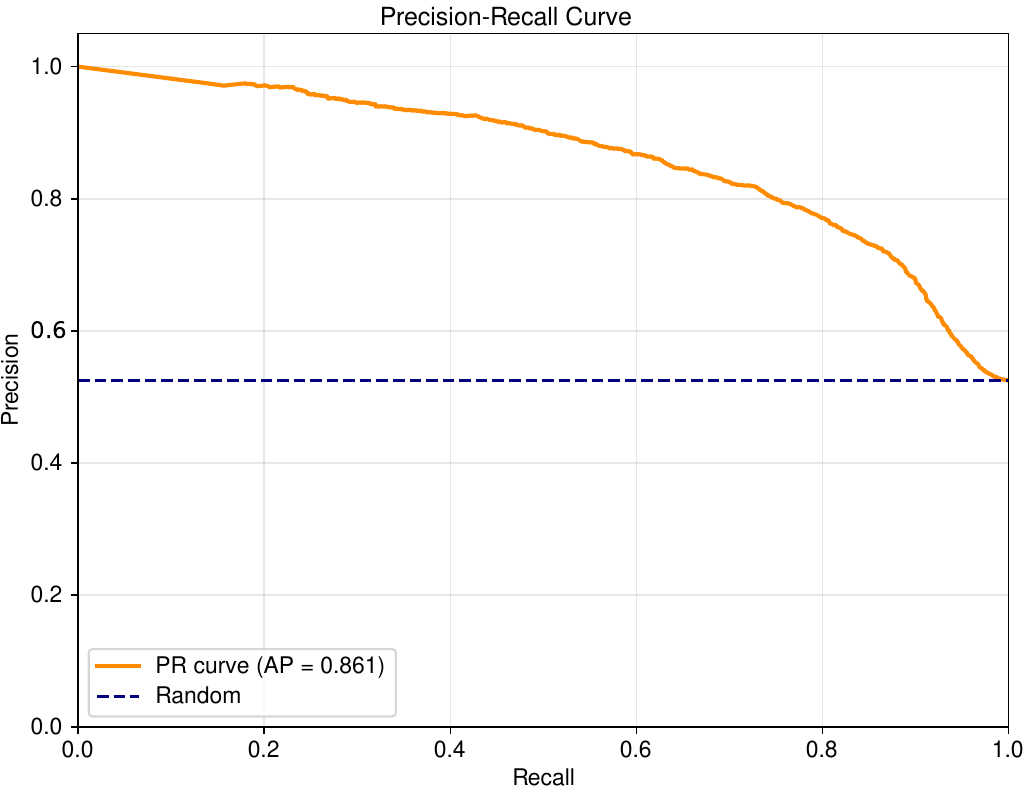}}
    \caption{\textbf{Molecule hold-out cold-start evaluation.} AUROC and AUPR performance on NPInter2 under unseen node settings.}
    \label{FIG:cold-start1}
\end{figure*}

\begin{table}[!t]
    \footnotesize
    \centering
    \caption{\textbf{Cold-start performance for RNA-protein interaction prediction on unseen nodes.} The best and second-best results are highlighted in \colorbox{lightgreen}{lightgreen} and \colorbox{lightred}{lightred}, respectively.}
    \label{tab:cold_nodes}
    \setlength{\tabcolsep}{12pt}
    \begin{tabular}{l|cc}
    \toprule[1.1pt]
    {Method} & {AUROC} & {AUPR} \\
    \midrule
    NPI-GNN~\cite{shen2021npi} & 0.511 & 0.520 \\
    IPMiner~\cite{pan2016ipminer} & 0.664 & 0.739 \\
    RPITER~\cite{peng2019rpiter} & 0.727 & 0.774 \\
    ZHMolGraph & \colorbox{lightred}{0.798} & \colorbox{lightred}{0.820} \\
    {EGRL(ours)} & \colorbox{lightgreen}{0.867} & \colorbox{lightgreen}{0.861} \\
    \bottomrule[1.1pt]
    \end{tabular}
\end{table}

\subsection{Answer to RQ4: Generalization capability validation}

Figure~\ref{FIG:cold-start1} presents the molecule hold-out cold-start results on the NPInter2 dataset. Evaluation on the unseen test set yields an AUROC of 0.867 and an AUPR of 0.861. Although these values are lower than those obtained under five-fold cross-validation (see \cref{Tab:other_models}), EGRL achieves state-of-the-art performance under the same evaluation protocol (see \cref{tab:cold_nodes}, with comparative data from~\cite{liu2025rna}). In particular, compared with the previous state-of-the-art method ZHMolGraph~\cite{liu2025rna} (AUROC: 0.798, AUPR: 0.820), EGRL improves AUROC by approximately 8.6\% and AUPR by 5.0\%.

\begin{figure*}[!t]
    \centering
    \includegraphics[width = 0.9\textwidth]{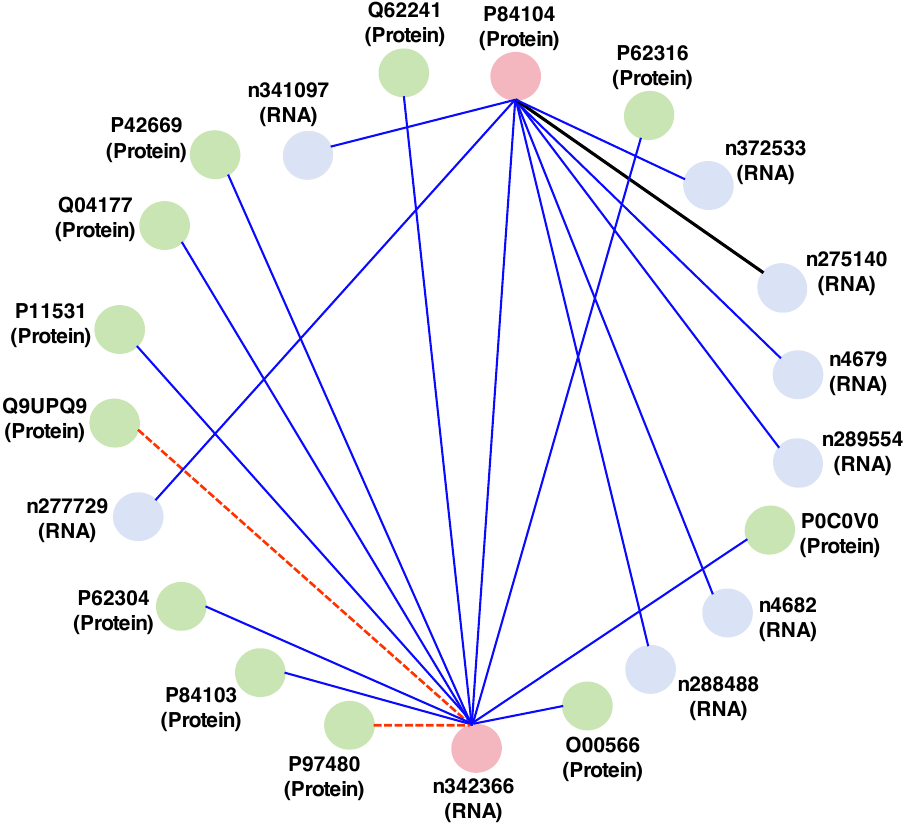}
    \caption{\textbf{Molecule hold-out interaction prediction.} Visualization of predicted interactions for removed molecules.}
    \label{FIG:molecule hold-out}
\end{figure*}

Figure~\ref{FIG:molecule hold-out} further illustrates the molecule hold-out results. After removing the protein (P84104) and the RNA (n342366) from NPInter2, these nodes are treated as cold-start targets. For protein P84104, only one interacting RNA is missed, while for RNA n342366, all interacting partners are correctly predicted. Some newly predicted interactions (red dashed lines) may correspond to false positives or potentially undiscovered interactions, indicating both prediction uncertainty and exploratory capability.

\begin{figure*}[!t]
    \centering
    \subfigure[AUROC of sequence similarity clustering validation.\label{AUC_NPInter}]{
    \includegraphics[width = 0.485\textwidth]{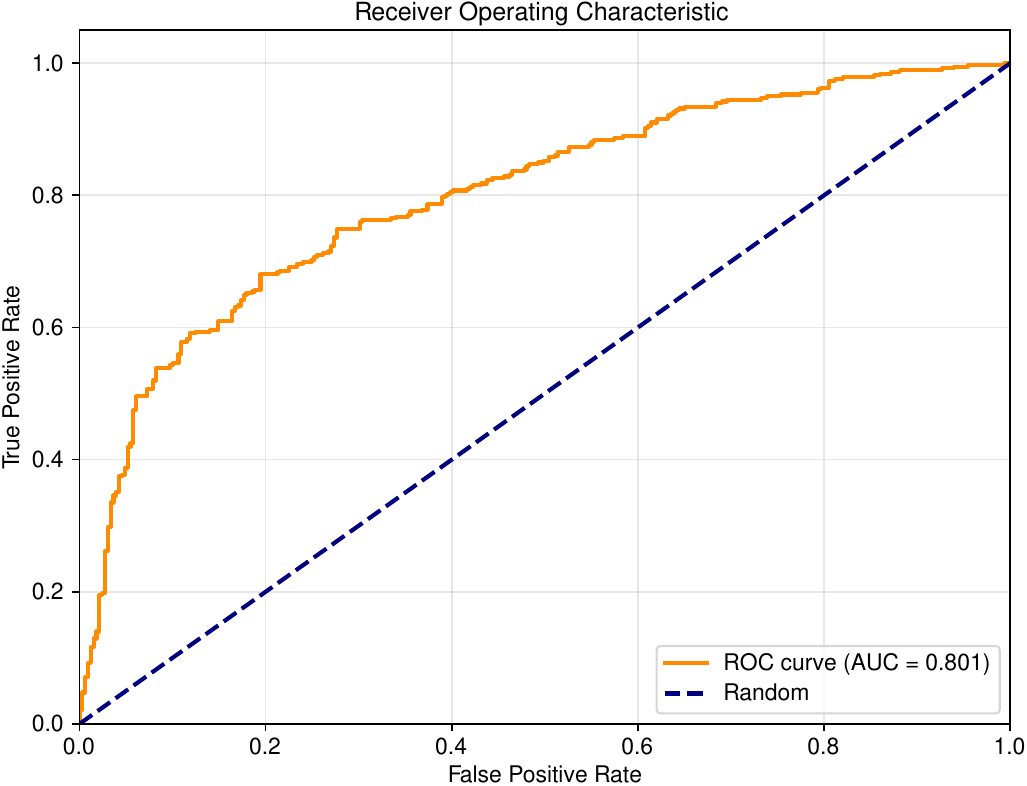}}
    \subfigure[AUPR of sequence similarity clustering validation.\label{AUPR_NPInter}]{
    \includegraphics[width = 0.485\textwidth]{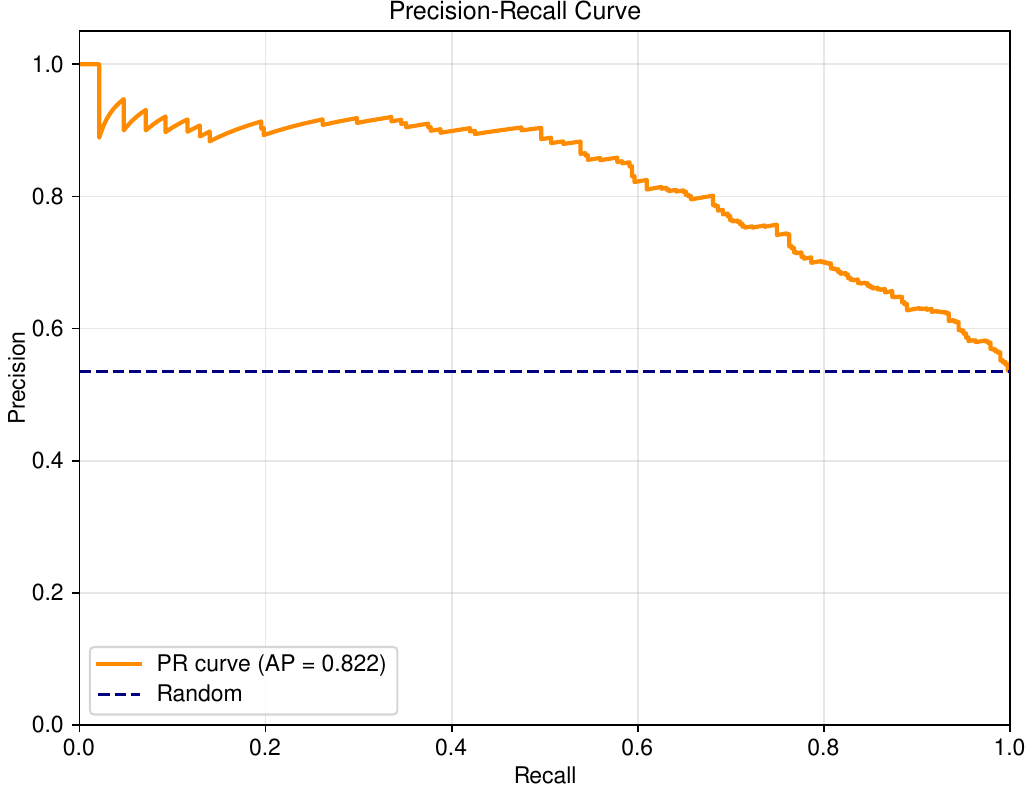}}
    \caption{Sequence-cluster-based cold-start evaluation. AUROC and AUPR performance under cluster-level data partitioning.}
    \label{FIG:cold-start2}
\end{figure*}

Figure~\ref{FIG:cold-start2} reports the results under the more stringent sequence-cluster-based setting. Evaluation on independent clusters yields an AUROC of 0.801 and an AUPR of 0.822. Compared with the molecule hold-out setting (AUROC: 0.867, AUPR: 0.861), both metrics decrease, likely because this strategy removes clusters with high sequence similarity to the training data, thereby reducing potential information leakage and increasing task difficulty.

\begin{figure*}[!t]
    \centering
    \includegraphics[width = 0.9\textwidth]{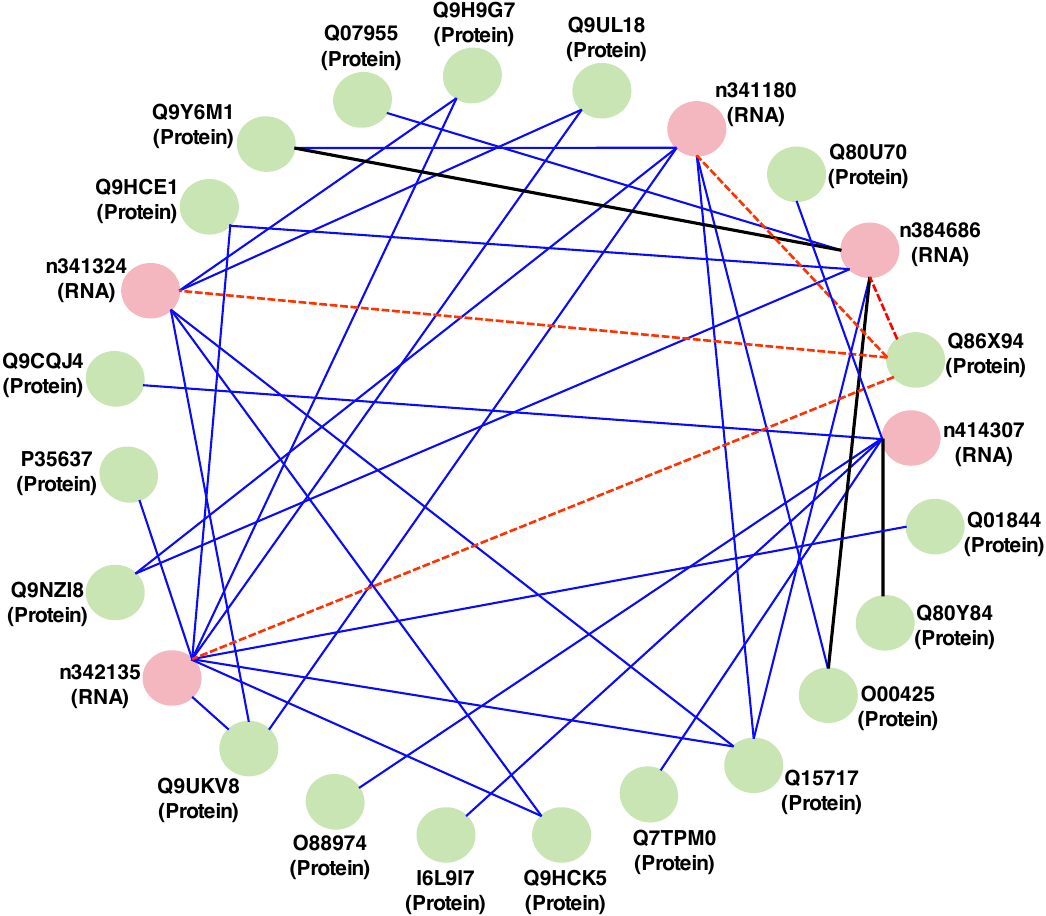}
    \caption{\textbf{Sequence-cluster-based interaction prediction.} Visualization of predicted interactions under cluster-level cold-start setting.}
    \label{FIG:Cold_study}
\end{figure*}

As shown in Figure~\ref{FIG:Cold_study}, although false negatives (black lines) and false positives (red lines) exist, the majority of interactions are correctly predicted (blue lines). This indicates that EGRL maintains stable predictive capability even under more challenging conditions.

\begin{tcolorbox}[colback = white,sharp corners,boxrule = 1pt]
    \textit{\textbf{Summary of answers to RQ4:}}
    Both validation settings demonstrate that EGRL achieves strong generalization in cold-start scenarios, maintaining reliable predictions even when specific molecules or entire sequence clusters are excluded from training.
\end{tcolorbox}

\section{Conclusions and future work}
\label{SEC:Conclusion and Future Work}

Graph neural networks (GNNs) have emerged as a promising paradigm for RNA-protein interaction (RPI) prediction. Given the multi-level and multi-type regulatory mechanisms underlying RPIs, constructing heterogeneous RNA-protein graphs that capture both node-level and path-level relational information is essential. Such modeling provides a solid foundation for advancing AI-driven biomedical research and drug discovery.

In this work, we propose EGRL for accurate RPI prediction. EGRL performs relation-aware encoding over multi-relational graphs. Its key strengths are threefold: (1) it automatically captures interaction patterns via implicit meta-path exploration; (2) it models multiple subgraphs (RNA-RNA, protein-protein, similarity, RPI, and generator-induced soft edges) using a multi-relational GAT and integrates their representations; and (3) it incorporates a graph generator trained jointly with the backbone to enhance generalization to unseen nodes (cold-start). Extensive experiments based on five-fold cross-validation demonstrate that EGRL achieves competitive performance across four benchmark datasets. Moreover, it shows robust generalization under both molecule hold-out and sequence-cluster-based settings, indicating its ability to capture underlying network connectivity and support the discovery of potential novel interactions.

For future work, we plan to explore multimodal data integration by incorporating additional biological information, such as RNA secondary structure, protein tertiary structure, gene expression profiles, and epigenetic signals, together with sequence features. This is expected to provide more comprehensive representations and further improve prediction accuracy and reliability. We also aim to develop more transferable frameworks for RPI prediction across species and tissue types, thereby extending the applicability of artificial intelligence in bioinformatics.

\section*{CRediT authorship contribution statement}

\textbf{Danyu Li:} Conceptualization, Data curation, Formal analysis, Methodology, Project administration, Visualization, Writing - original draft;
\textbf{Ling Zhou:} Conceptualization, Formal analysis, Methodology, Validation, Writing - original draft, Writing - review \& editing;
\textbf{Rubing Huang:} Funding acquisition, Methodology, Supervision, Validation, Writing - original draft, Writing - review \& editing;
\textbf{Xian Zhong:} Formal analysis, Methodology, Validation, Visualization, Writing - review \& editing;
\textbf{Bin Zou:} Conceptualization, Data curation, Validation, Writing - review \& editing;
\textbf{Kui Jiang:} Conceptualization, Formal analysis, Methodology, Writing - review \& editing.

\section*{Declaration of Generative AI and AI-assisted technologies in thewriting process}

The authors used a generative AI tool solely for language editing, including grammar, formatting, and spelling checks. The authors take full responsibility for the content of the manuscript.

\section*{Declaration of competing interest}

The authors declare that they have no known competing financial interests or personal relationships that could have appeared to influence the work reported in this paper.

\section*{Acknowledgments}

This work was supported in part by the Science and Technology Development Fund of Macau, Macao SAR, under Grants 0069/2025/RIB2 and 0021/2023/RIA1, the National Natural Science Foundation of China under Grant 62271361, the Hubei Provincial Key Research and Development Program under Grant 2024BAB039, and the Fundamental Research Funds for the Central Universities under Grant 104972026KFYzxk0029.

\section*{Data availability}

Data will be made available on request.

\bibliographystyle{elsarticle-num}
\bibliography{EGRL}

\end{document}